\pdfoutput=1

\documentclass[11pt]{article}

\usepackage[]{EMNLP2022}
\usepackage{stfloats}

\usepackage{times}
\usepackage{latexsym}
\usepackage[T1]{fontenc}
\usepackage{graphicx}

\usepackage[utf8]{inputenc}

\usepackage{microtype}

\usepackage{hyperref}       
\usepackage{url}            
\usepackage{booktabs}       
\usepackage{amsfonts}       
\usepackage{nicefrac}       
\usepackage{microtype}      
\usepackage{xcolor}         
\usepackage{inconsolata}
\usepackage{algorithm}
\usepackage{graphicx}
\usepackage{bm}
\usepackage{bbm}
\usepackage{gensymb}
\usepackage{color}
\usepackage{mathtools}
\usepackage{subfigure}
\usepackage{marvosym}
\renewcommand{\thefootnote}{}

\newcommand {\ourmethod}[1]{LVP-M$^{3}$}

\title{LVP-M$^{3}$: Language-aware Visual Prompt for Multilingual \\Multimodal Machine Translation}
\author{Hongcheng Guo*$^{1}$, Jiaheng Liu*$^{1}$, Haoyang Huang$^{2}$, Jian Yang$^{1}$,\\\textbf{Zhoujun Li\textsuperscript{\Letter}}$^{1}$, \textbf{Dongdong Zhang}$^{2}$ , \textbf{Zheng Cui}$^{2}$, \textbf{Furu Wei}$^{2}$\\
       $^{1}$Beihang University \\
       $^{2}$Microsoft Research Asia \\
        \texttt{\{hongchengguo,liujiaheng,jiaya,lizj\}@buaa.edu.cn}\\\texttt{\{haohua,dozhang,zhcui,fuwei\}@microsoft.com}}
     
\begin{document}
\maketitle
\let\thefootnote\relax\footnotetext{* First two authors contributed equally.}
\let\thefootnote\relax\footnotetext{\textsuperscript{\Letter} Corresponding author.}
\begin{abstract}
Multimodal Machine Translation (MMT) focuses on enhancing text-only translation with visual features, which has attracted considerable attention from both natural language processing and computer vision communities. Recent advances still struggle to train a separate model for each language pair, which is costly and unaffordable when the number of languages increases in the real world. In other words, the multilingual multimodal machine translation (\textbf{Multilingual MMT}) task has not been investigated, which aims to handle the aforementioned issues by providing a shared semantic space for multiple languages. Besides, the image modality has no language boundaries, which is superior to bridging the semantic gap between languages. To this end,
we first propose the Multilingual MMT task by establishing two new Multilingual MMT benchmark datasets covering seven languages.
Then, an effective baseline LVP-M$^{3}$ using visual prompts is proposed to support translations between different languages,
which includes three stages (token encoding, language-aware visual prompt generation, and language translation). Extensive experimental results on our constructed benchmark datasets demonstrate the effectiveness of LVP-M$^{3}$ method for Multilingual MMT.
\end{abstract}

\section{Introduction}
Multimodal Machine Translation (\textbf{MMT}) extends the conventional text-based machine translation by taking corresponding images as additional inputs~\cite{lin2020dynamic,vision_features} to mitigate the problems of data sparsity and ambiguity~\cite{ive-etal-2019-distilling,um4} when compared with purely text-based machine translation. Similar to other multimodal tasks (e.g.,  
visual question answering~\cite{antol2015vqa,shih2016look},
 image captioning~\cite{vinyals2015show,jia2015guiding} and
video-text retrieval~\cite{liu2022cross}), MMT aims to exploit the effectiveness of vision information for the machine translation task.

\begin{figure}[t]
\begin{center}
\includegraphics[width=1.0\linewidth]{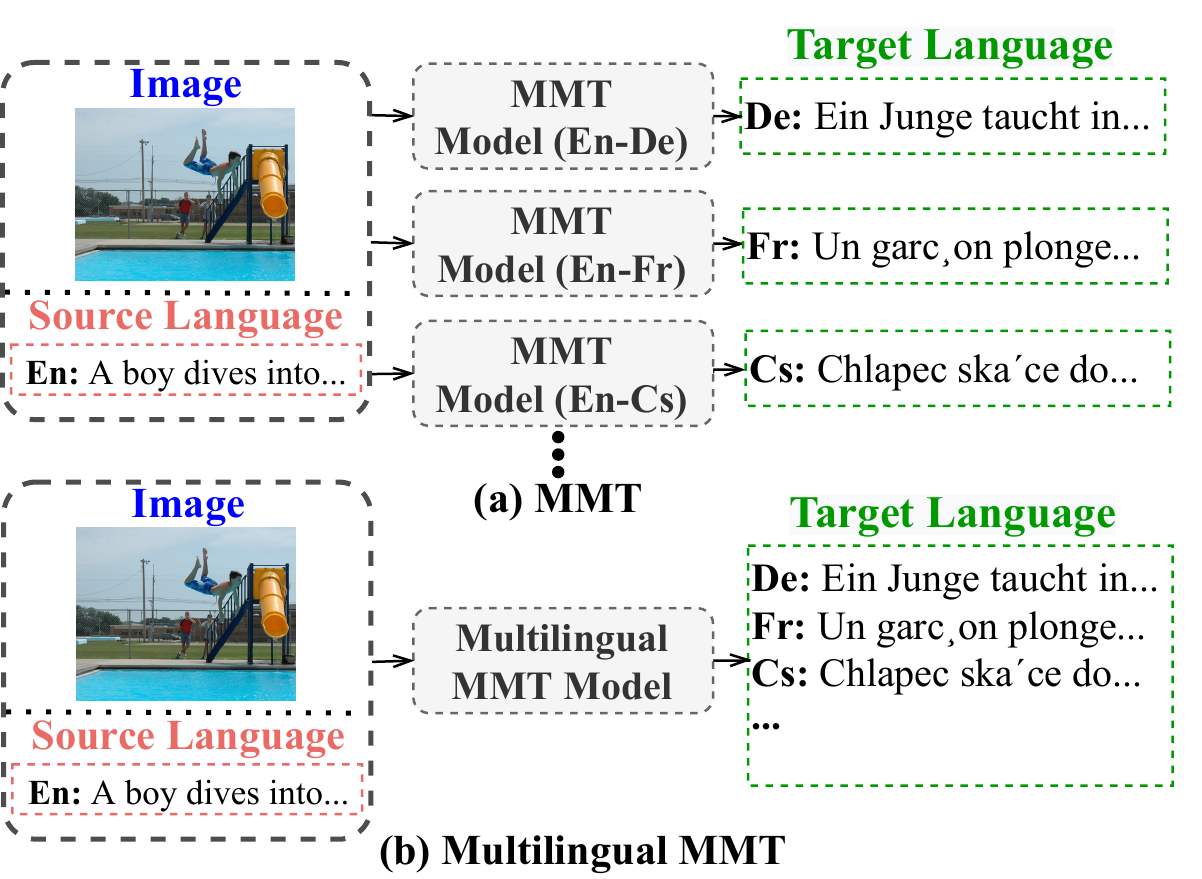}
\caption{Comparison of MMT and Multilingual MMT. (a) For MMT, we need to train different MMT models to support translations between different language pairs (e.g., ``En-De'' represents to translate the English to German).
(b). For Multilingual MMT, we only need one single model to translate the source language to different target languages.}
\label{fig:intro}
\end{center}
\end{figure}
Moreover, MMT has broad applications~\cite{zhou2018visual}, such as multimedia
news and movie subtitles in different languages.

However,
as shown in Fig.~\ref{fig:intro}(a),
previous MMT models (e.g., DCCN~\cite{lin2020dynamic}) can
handle a single language translation pair (e.g.,
English $\rightarrow$ German, English $\rightarrow$ French) well,
but training a separate model for each language pair is unaffordable considering there are thousands of languages in the world.
A straightforward solution to reduce computational cost is to use one model for handling the translations of multiple languages as shown in Fig.~\ref{fig:intro}(b).
Meanwhile, multilingual machine translation has been investigated for many years~\cite{conneau2020unsupervised},
but these existing methods only consider the language as the input, where the vision context has been ignored. Therefore, in our work, we first propose the Multilingual Multimodal Machine Translation  (\textbf{Multilingual MMT}) task to achieve the translations for multiple languages using one single model.

To eliminate the above limitations, we propose a simple and effective \textbf{LVP-M$^{3}$} method, including Token Encoding, Language-aware Visual Prompt Generation (LVPG), and Language Translation.
Specifically, in the token encoding stage, we use the pre-trained vision encoder to extract the visual tokens. Then, we follow~\cite{johnson2017google} to utilize the Transformer to encode the textual tokens. In LVPG, inspired by \cite{yang2019condconv} and \cite{tian2020conditional}, a controller network in Fig.~\ref{fig:pipeline} is leveraged to dynamically generate the parameters of the mapping network conditioned on the target language. Further, the mapping network outputs the language-aware visual prompts. After that, during the language translation, following the works (e.g., ViLBERT~\cite{lu2019vilbert}), we utilize co-Transformer to generate the vision-guided language tokens. Then the Transformer decoder is adopted to predict the translation results.

Extensive experiments are conducted on our proposed benchmark datasets for {LVP-M$^{3}$}. Results show that our model achieves the state-of-the-art performance in all translation directions, especially outperforming the text-only multilingual model by $4.3$ BLEU scores on average.

The contributions of this work are summarized as follows:
\begin{itemize}
    \item
    We first propose the Multilingual Multimodal Machine Translation (Multilingual MMT) to handle the translations for multiple language pairs,
    which investigates the effect of vision modality for multilingual translation and reduces the computation costs of existing MMT methods for multiple languages.
    \item For Multilingual MMT, we propose an effective language-aware visual prompt generation strategy to produce different visual prompts for different target languages based on the vision modality and type of the target language.
    \item We establish two Multilingual MMT benchmark datasets to nourish the further research on Multilingual MMT, and extensive experiments on these datasets demonstrate the effectiveness of our proposed \ourmethod{} method.
\end{itemize}

\section{Related Works}
\paragraph{Multimodal Machine Translation.}
The multimodal context plays a key role in Multimodal Machine Translation (MMT). Recent MMT methods can be divided into three categories: (1) Using global visual features directly \cite{calixto2017incorporating}.
For instance, ~\citet{huang2016attention} proposes to concatenate global and regional visual features with source sequences.
(2) Exploiting visual features via attention scheme~\cite{libovicky2017attention,helcl2018cuni}.
~\citet{calixto2017doubly} introduces the visual features into the MMT model by using an independent attention module. 
(3) Combining other vision tasks with the translation task by multitask learning \cite{calixto2019latent,yin2020novel}.
~\citet{elliott2017imagination} decomposes multimodal translation into two sub-tasks (i.e., translation and visual grounding).
Recently,
~\cite{huang-etal-2020-unsupervised-multimodal} focuses on unsupervised setting for MMT, which utilizes pseudo visual pivoting and visual content to improve the cross-lingual alignments in latent space. In contrast, ~\ourmethod{} considers fully-supervised multilingual setting by mapping vision embeddings into different feature spaces and achieving the purpose of using one MT model for handling translations of multiple languages.
Besides, reducing computation cost is vital for many tasks~\cite{9219251,liu2022coupleface,9844448} and we focus on the Multilingual MMT task by using one single model for efficiency.

\noindent\textbf{Multilingual Language Models}. 
Pre-trained multilingual Transformer-based language models (e.g., mBERT~\cite{kenton2019bert}
 and XLM-R~\cite{conneau2020unsupervised}) utilize the same pre-training strategies as their respective monolingual counterparts (e.g., BERT~\cite{kenton2019bert} and RoBERTa~\cite{liu2019roberta}).
They are pre-trained via the masked language
modeling objective (MLM) Strategy.
~\citet{artetxe2020cross} proposes a method to transfer monolingual representations to new languages in an unsupervised fashion and provide new insights into the generalization abilities of multilingual models.
~\citet{hu2020xtreme} introduces the 
Cross-lingual Transfer Evaluation of Multilingual Encoders 
(XTREME) benchmark to evaluate the
cross-lingual generalization capabilities,
~\citet{karthikeyan2019cross} also provides a comprehensive study of the contribution of different components in M-BERT to its cross-lingual ability.~\citet{rust2021good} shows that monolingually adapted tokenizers can robustly improve the monolingual performance
of multilingual models.
Overall, when compared with these methods, we focus on the multilingual setting for MMT,
which has not been investigated before.

\noindent\textbf{Vision-Language Models}. 
The success of vision-language models can be credited to the following three important reasons: 
Transformers~\cite{9854132,vaswani2017attention}, contrastive representation learning~\cite{radford2021learning,li2020unicoder}, and large-scale training datasets~\cite{sharma2018conceptual,Miech2019HowTo100MLA}.
Previous Transformer-based multimodal models (~\cite{tan2019lxmert,chen2020uniter,gan2020large,bugliarello2021multimodal})
jointly encode text tokens and image region features by preprocessing images using object detection models. 
The image region features are projected into the joint embedding space of the multimodal Transformer, and then the multi-head attention attends to all text and image inputs to learn a joint representation of both modalities. Besides, ~\citet{kamath2021mdetr} avoids using object detectors as a black box for pre-extracting these region features and incorporates the object detector end-to-end with the multimodal Transformer to achieve flexibility and better representation capacity.
Recently,
a representative approach CLIP~\cite{radford2021learning} is proposed, which trains two neural network-based encoders using a contrastive loss to match pairs of images and texts. After consuming 400 million data pairs, the CLIP model demonstrates a remarkable zero-shot image recognition capability, and has been applied to many downstream tasks.
 For example, ~\citet{shen2022how} proposes to leverage the CLIP model for different vision-language models across various tasks (e.g., image caption, visual question answering).
 In our work, we aim to investigate the effectiveness of the multimodal information for Multilingual MMT.

\section{Datasets}
We introduce two Multilingual MMT benchmark datasets (i.e., M$^{3}$-Multi30K, M$^{3}$-AmbigCaps) using Multi30K~\cite{elliott-etal-2016-multi30k} and  AmbigCaps~\cite{vision_matters}.
Here,
we descried the details of the M$^{3}$-Multi30K and  M$^{3}$-AmbigCaps.
\begin{figure}[!htp]
\begin{center}
\includegraphics[width=1.0\linewidth]{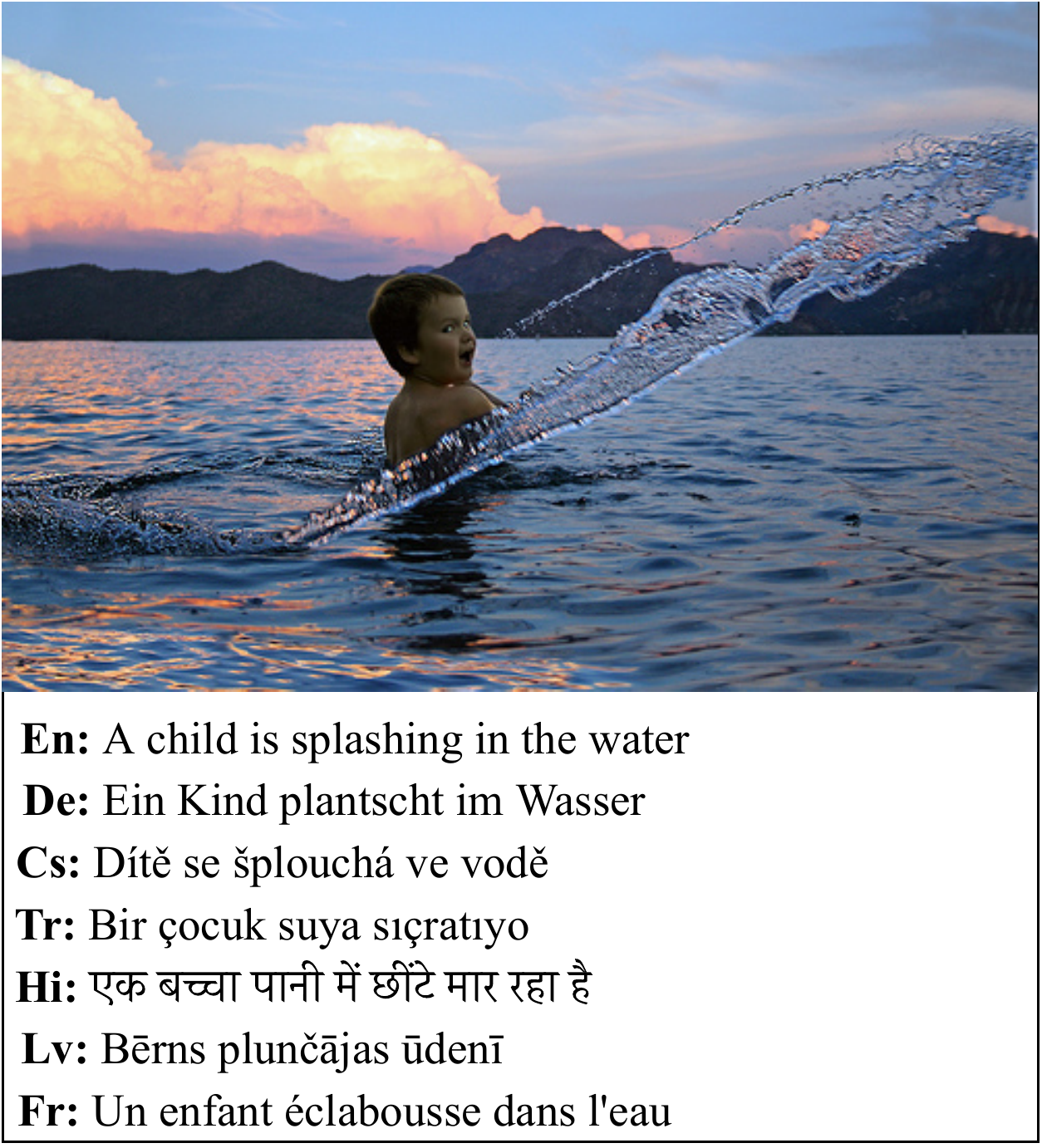}
\caption{Example of an image with its  descriptions of seven different languages.}
\label{fig:sample}
\end{center}
\end{figure}

\noindent\textbf{Data Construction.}
The widely-used Multi30K dataset for the MMT task is based on the Flickr30K Entities dataset~\cite{flickrentitiesijcv}.
For each image of Multi30K, one of the English (En) descriptions is selected in Flickr30K Entities.
Currently, the English description of each image is translated into German (De), French (Fr), and Czech (Cs)~\cite{elliott-etal-2017-findings,barrault-etal-2018-findings}.
To support more languages from different language families and various language distributions for Multilingual MMT,
 we extend the existing Multi30K dataset with additional three languages as shown in Table~\ref{Tab:lang},
 where one sample of the M$^{3}$-Multi30K dataset is shown in Fig.~\ref{fig:sample}.
 
\begin{table}[t]
    \centering

    \begin{tabular}{c|c|c|c}
    \toprule
    Language  & ISO & Family&Speakers\\ 
    \hline
English &En &Germanic &400M\\
German &De &Germanic&95M\\
French &Fr &Romance  &250M\\
Czech &Cs &Slavic  &11M\\
Hindi &Hi &Indo-Aryan &800M\\
Turkish &Tr &Turkic &65M\\
Latvian &Lv&Baltic &2M\\
    \bottomrule
    \end{tabular}
        \caption{Languages covered by our proposed M$^{3}$-Multi30K and M$^{3}$-AmbigCaps datasets.}
            \label{Tab:lang}

\end{table}
Specifically,
in the annotation process,
based on the recent state-of-the-art multilingual machine translation model XLM-R~\cite{conneau2020unsupervised},
we first translate the English description into Hindi (Hi), Turkish (Tr), and Latvian (Lv) for each image in Multi30K.
Then,
we hire independent native speakers to verify
and improve the quality of the translation results of different languages.
In addition,
as the original AmbigCaps~\cite{vision_matters} dataset only contains two types of languages,
we use a similar way to extend AmbigCaps into additional five languages in M$^{3}$-AmbigCaps.

\noindent\textbf{Data Splits.}
In M$^{3}$-Multi30K,
the number of image-translation pairs for training and testing data are {29000, 1000}, respectively.
In  M$^{3}$-AmbigCaps,
the number of image-translation pairs for training and testing data are {89600, 1000}, respectively.
We will released these datasets.


\section{Method}
\subsection{Multilingual MMT}
Supposing we have $M$ languages $\{L_{m}\}_{m=1}^{M}$ and $N$ bilingual corpora $\{D_{n}\}_{n=1}^{N}$ under the multilingual setting,
the dataset $D_{n}$ consists of $K$ parallel sentences $\{(x_{L_i}^{k}, x_{L_j}^{k})\}_{k=1}^{K}$ between language $L_i$ and $L_j$, where $K$ is the number of training instances and each instance has the corresponding image $z_k$.
Given the corpora, we can train a Multilingual MMT model that enables the translation among different languages with the help of image modality. The training objective of the Multilingual MMT is learnt with a combination of different languages:
\begin{equation}
    \mathcal{L}_{mt} =-\sum_{i,j,k}{\mathrm{1og} \mathbf{P}(x_{L_i}^{k}; x_{L_j}^{k}; z_k)},
    \label{eq1}
\end{equation}where the Multilingual MMT model uses a complete shared model for all translation directions. 
In this work, we adopt Transformer as the backbone model for language encoding and pre-trained vision branch of the CLIP model~\cite{radford2021learning} for vision modality. A target language token 
$t_{L_{j}}$ is prefixed to each source sentence to indicate the translation direction~\cite{johnson2017google}.

\begin{figure*}[!htp]
\begin{center}
\includegraphics[width=1.0\linewidth]{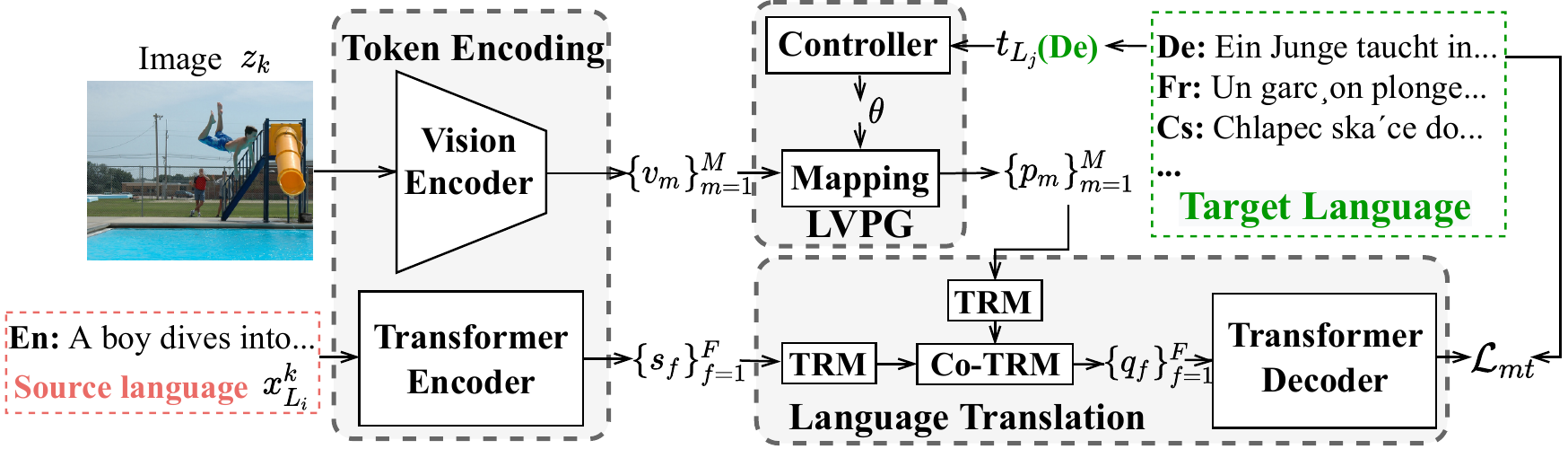}
\caption{The overall framework of our proposed LVP-M$^{3}$ method for Multilingual MMT task,
which includes three stages (i.e., token encoding and language-ware visual prompt generation (LVPG) and language translation). Here, we take an example by translating English (En) to German (De). ``TRM'' and ``Co-TRM'' represent the Transformer and co-Transformer models, respectively.}
\label{fig:pipeline}
\end{center}
\end{figure*}
\subsection{LVP-M$^{3}$}
As shown in Fig.~\ref{fig:pipeline},
our proposed LVP-M$^{3}$ method includes three stages: token encoding,
language-aware visual prompt generation and language translation.
Specifically,
in training,
give each image $z_k$, the parallel sentences $\{(x_{L_i}^{k}; x_{L_j}^{k})\}$ from source language $L_i$ and target language $L_j$,
and the target language token embedding $t_{L_j}$,
in token encoding stage,
we first use the vision encoder to extract the visual token features $\{v_{m}\}_{m=1}^{M}$  based on $z_k$,
where $M$ is the number of visual tokens.
Then,
we utilize the Transformer encoder to extract the source language tokens $\{s_{f}\}_{f=1}^{F}$,
where $F$ is the number of language tokens.
In language-aware visual prompt generation (LVPG) stage,
we  map the $\{v_{m}\}_{m=1}^{M}$ into the language-aware visual prompt $\{p_{m}\}_{m=1}^{M}$ conditioned on $t_{L_j}$ to generate different visual prompts for different target languages,
where we propose to adopt the controller network to dynamically generate the parameters of the mapping network.
In language translation stage,
we first adopt the co-attention strategy to generate the vision-guided  language tokens $\{q_{f}\}_{f=1}^{F}$ based on  $\{p_{m}\}_{m=1}^{M}$  and $\{s_{f}\}_{f=1}^{F}$.
Then,
we use the $\{q_{f}\}_{f=1}^{F}$ as the input of the Transformer decoder to predict the translation results and compute the loss in Eq.~\ref{eq1} using the predicted translation results and the ground-truth target language $x_{L_j}^{k}$.

\subsubsection{Token Encoding}
For each image $z_k$,
we directly use the vision backbone (e.g., the pre-trained vision branch of the widely-used CLIP model~\cite{radford2021learning}) as the vision encoder to extract the visual tokens for $z_k$ as follows:
\begin{equation}
    \{v_{m}\}_{m=1}^{M}= \mathcal{H}(z_k),
\end{equation}
where $\mathcal{H}$ denotes the vision encoder and $M$ is the number of visual tokens.

Similarly,
given the source language $x_{L_i}^{k}$,
based on the Transformer encoder $ \mathcal{E}$,
the source language tokens $\{s_{f}\}_{f=1}^{F}$ are extracted as follows:
\begin{equation}
    \{s_{f}\}_{f=1}^{F}= \mathcal{E}(x_{L_i}^{k}),
\end{equation}
where $F$ is defined as the number of source language tokens.





\subsubsection{Language-aware Visual Prompt Generation}
In language-aware visual prompt generation stage of Fig.~\ref{fig:pipeline},
motivated by 
recent works (e.g., dynamic filter networks~\cite{jia2016dynamic}
and CondConv~\cite{yang2019condconv}) based on conditional convolutions,
where the filters of conditional convolutions are conditioned on the input and are dynamically generated by another network to improve the capacity of the neural network,
 we extend this idea to generate the visual prompt conditioned on the target language type $t_{L_j}$ (e.g., German) to map the extracted the visual tokens into different embedding spaces for different  target language.
Specifically,
in Fig.~\ref{fig:pipeline},
based on the embedding of the target language token $t_{L_j}$,
we utilize a controller network  $\mathcal{C}$ implemented by two fully-connected layers with ReLU~\cite{nair2010rectified} activation function to generate the parameters $\theta$ of the mapping network $\mathcal{M}$ as follows:
\begin{equation}
    \theta  = \mathcal{C}(t_{L_j}).
    \label{theta}
\end{equation}
After that,
we generate the language-aware visual prompt $\{p_{m}\}_{m=1}^{M}$ as follows:
\begin{equation}
    \{p_{m}\}_{m=1}^{M} = \mathcal{M}(\{v_{m}\}_{m=1}^{M}, \theta).
\end{equation}
$\theta$ is the generated parameters in Eq.~\ref{theta},
which is assigned to the mapping network $\mathcal{M}$.
In this way,
when translating source language into different target languages,
the $\theta$ will be generated according to type of target language tokens, and the visual tokens $\{v_{m}\}_{m=1}^{M}$ can be mapped into different visual prompts according to the type of the target language.


\subsubsection{Language Translation}
In Fig.~\ref{fig:pipeline},
based on the  source language tokens $\{s_{f}\}_{f=1}^{F}$ and language-aware visual prompt $\{p_{m}\}_{m=1}^{M}$,
we first generate the vision-guided source language tokens based on co-attention strategy,
which are widely used for fusing the information from another modality in vision-language models~\cite{lu2019vilbert}.
Then, we predict the translation results using the Transformer decoder.

Specifically,
we utilize the Transformer module implemented by self-attention to fuse the information from other tokens within each modality for $\{s_{f}\}_{f=1}^{F}$ and   $\{p_{m}\}_{m=1}^{M}$,
respectively,
and we represent the updated source language tokens and visual prompt as $\mathbf{S}$ and $\mathbf{P}$,
respectively.
Then,
we take $\mathbf{S}$ as the query,
and the $\mathbf{P}$ as the key and value in the co-attention module to generate the  vision-guided source language tokens $\{q_{f}\}_{f=1}^{F}$
as follows:
\begin{equation}
\{q_{f}\}_{f=1}^{F} = \overset{H}{\underset{h=1}{\big\|}}  \mathtt{SF} \left (\frac{\boldsymbol\phi^h_Q(\mathbf{S}) \boldsymbol\phi^h_K(\mathbf{P})^\top}{\sqrt{C}}\right) \boldsymbol\phi_V^h (\mathbf{P}), \label{eq:co-attention}
\end{equation}
where $\|_{h=1}^H$ is the concatenation of the $H$ attentive features along the channel dimension. $\mathtt{SF}$ represents the softmax operation. 
$\boldsymbol\phi_Q^h(\cdot), \boldsymbol\phi_K^h(\cdot)$ and $ \boldsymbol\phi_V^h(\cdot)$ are the corresponding linear projection operations of the $h$-th head  for the query, the key and the value, respectively. 
$C$ denotes the number of feature channels.
After the operation of Eq.~\ref{eq:co-attention},
 other operations (e.g., FFN, layer normalization~\cite{ba2016layer}) of standard attention scheme~\cite{vaswani2017attention}
are used.


Finally,
at inference,
based on  $\{q_{f}\}_{f=1}^{F}$,
we use the Transformer decoder to predict the target language sequence in our \ourmethod{}.
\section{Experiments}
We evaluate our proposed \ourmethod{} method on the multilingual dataset including
7 languages and {6 translation directions.} 
In all experiments, English (En) is treated as the pivot language for Multilingual MMT setting.

\begin{table*}[t]
\centering
\resizebox{1.0\textwidth}{!}{
\begin{tabular}{l|cccccc|c}
\toprule
Model (En$\rightarrow$X)   & Fr & Cs & De & Lv & Hi & Tr & Avg$_{all}$ \\
\midrule
\multicolumn{8}{c}{\textit{Text-only Multilingual MT Systems}} \\ 
\midrule
  Text Transformer \cite{english-centered}   & 61.8 & 32.8 & 40.6 & 51.2 & 59.0 & 53.8 & 49.8 \\
\midrule
	\multicolumn{8}{c}{\textit{Multilingual MMT Systems}} \\
\midrule

   Vision Matters (Gated fusion) \cite{vision_matters}  & 62.5 & 32.9 & 41.2 & 52.1 & 59.6 & 54.2 & 50.4 \\
   Vision Matters (Concatenation) \cite{vision_matters}  &59.7  & 33.1 & 39.8 & 50.3 & 57.6 & 51.4 & 48.6 \\
  \ourmethod{} (w/o LVPG)  & 62.2 & 33.4 & 40.9 & 51.6 & 59.3 & 54.0 & 50.2\\
  \bf \ourmethod{} (Our method)  & \bf 63.7 & \bf 34.6  & \bf 43.2 & \bf 53.5 & \bf 61.4 & \bf 55.6 & \bf 52.0  \\
\bottomrule
\end{tabular}}
\caption{The BLEU scores of different methods on M$^{3}$-Multi30K test set. Five multilingual baselines are compared by us. The bottom part shows the results of the multilingual models trained with text and vision modalities. The best results are highlighted.}
\label{table:multi30k}
\end{table*}

\begin{table*}[t]
\centering
\resizebox{1.0\textwidth}{!}{
\begin{tabular}{l|cccccc|c}
\toprule
Model (En$\rightarrow$X)  & Fr & Cs & De & Lv & Hi & Tr & Avg$_{all}$ \\
\midrule
\multicolumn{8}{c}{\textit{Text-only Multilingual MT Systems}} \\ 
\midrule
  Text Transformer \cite{english-centered}   & 62.3 & 47.8 & 49.0 & 46.6 & 52.4 & 35.9 & 49.0 \\
\midrule
	\multicolumn{8}{c}{\textit{Multilingual MMT Systems}} \\
\midrule
  Vision Matters (Gated fusion) \cite{vision_matters}  & 64.3 & 50.3 & 51.2 & 48.5 & 54.1 & 38.7 & 51.2 \\
  Vision Matters (Concatenation) \cite{vision_matters}  &62.6  & 47.6 & 48.7 & 45.9 & 52.7 & 36.0 & 48.9 \\
  \ourmethod{} (w/o LVPG) & 63.4 & 49.2 & 50.3 & 47.9 & 52.4 & 37.1 & 50.1\\
  \bf \ourmethod{} (Our method) & \bf 65.7 & \bf 52.9  & \bf 53.7 & \bf 51.6 & \bf 56.3 & \bf 42.7 & \bf 53.8  \\
\bottomrule
\end{tabular}}
\caption{The BLEU scores of different methods on M$^{3}$-AmbigCaps test set. Five multilingual baselines are compared by us. The bottom part shows the results of the multilingual models trained with text and vision modalities.The best results are highlighted.}
\label{table:amcaps}
\end{table*}
\subsection{Experimental Setting}
\noindent\textbf{Implementation Details.}
Our implementation is based on the Fairseq \cite{ott2019fairseq} toolbox.
 We utilize Sentencepiece tokenizer.
The model in Fig.~\ref{fig:pipeline} consists of $6$ Transformer encoder/decoder layers. The number of attention heads in all Transformer layers is set as $12$.
For training, we take the Adam optimizer \cite{kingma2014adam} with $\beta_1=0.9$ and $\beta_2=0.98$. The learning rate warms up from 1e-7 to 1e-4 in $2000$ steps and then decays based on the inverse square root of the update number. The maximum number of tokens in each mini-batch is $4096$. Dropout and label-smoothing rate are set as $0.3$ and $0.1$, respectively.
For vision encoder, by default, we adopt the vision branch of CLIP based on the ViT-L/14 model. The effect of different vision backbones is discussed in our ablation study. All models are trained for $30$ epochs and evaluated on one single linux machine with 8 NVIDIA A100 GPUs (80G).

\noindent\textbf{Evaluation.}
We compute the cumulative 4-gram BLEU scores to evaluate the quality of translation. During inference, the beam search strategy is performed with a beam size of $5$ for the target sentence generation. We set the length penalty as $1.0$.

\noindent\textbf{Baseline Methods.}
As we are the first multilingual method in this area, we reproduce methods including Text Transformer~\cite{english-centered}, the Vision Matters (Gated fusion) ~\cite{vision_matters}, and the Vision Matters (Concatenation)~\cite{vision_matters} in the multilingual translation setting for a fair comparison.
Besides, we also report the results of \ourmethod{} (w/o LVPG), where we directly adopt the co-attention strategy in~\citet{lu2019vilbert} to generate the vision-guided language tokens using the source tokens with the visual features. 

\subsection{Results on M$^{3}$-Multi30K}
To demonstrate the effectiveness of \ourmethod{},
we compare our method with baseline methods on M$^{3}$-Multi30K under the multilingual MMT setting in Table~\ref{table:multi30k}.
It should be mentioned that the Vision Matters (Gated fusion) and the Vision Matters (Concatenation) are originally proposed in the bilingual setting,
and we reproduce these methods in the multilingual setting for a fair comparison. 
In Table~\ref{table:multi30k}, our \ourmethod{} achieves the best BLEU scores in all translation directions.
Specifically,
first, when compared with text Transformer with only text information, \ourmethod{} outperforms by $+2.2$ BLEU scores on average,
which demonstrates the effectiveness of visual context for Multilingual MMT. 
Second, when compared with baseline method \ourmethod{} (w/o LVPG),
\ourmethod{} also achieves better performance on all settings, which verifies the effectiveness of our proposed language-aware prompt generation module for Multilingual MMT. Among all translation directions, the task of En$\to$De achieves the most improvement. Because English and German are from the same language family, both languages can share the similar semantic knowledge by cross-lingual transfer. 


\begin{table*}[!htp]
\centering
\begin{tabular}{l|cccccc|c}
\toprule
Model (En$\rightarrow$X)  & Fr & Cs & De & Lv & Hi & Tr & Avg$_{all}$ \\
\midrule
   \ourmethod{} (Static)  & 62.0  & 33.1 & 41.1 & 51.7 & 59.6 & 54.2 & 50.3 \\
  \bf \ourmethod{} (LVPG) & \bf 63.7 & \bf 34.6  & \bf 43.2 & \bf 53.5 & \bf 61.4 & \bf 55.6 & \bf 52.0   \\
\bottomrule
\end{tabular}
\caption{Comparison of different vision prompt generation methods with BLEU scores.}
\label{table:language-aware_generation}
\end{table*}

\begin{table*}[!htp]
\centering
\begin{tabular}{l|cccccc|c}
\toprule
Model (En$\rightarrow$X)  & Fr & Cs & De & Lv & Hi & Tr & Avg$_{all}$ \\
\midrule
 \ourmethod{}+ResNet50   & 62.3 & 33.3 & 41.7 & 52.3 & 61.1 & 54.0   & 50.8 \\
 \ourmethod{}+ResNet101   & 62.8 & 33.8 & 42.1 & 52.5 & 60.7 & 54.2    & 51.1 \\
    \ourmethod{}+ViT-L/14  & \bf 63.7 & \bf 34.6  & \bf 43.2 & \bf 53.5 & \bf 61.4 & \bf 55.6 & \bf 52.0 \\
\bottomrule
\end{tabular}
\caption{Comparison different visual backbones with BLEU scores.}
\label{table:backbone}
\end{table*}

\subsection{Results on M$^{3}$-AmbigCaps}
Results of M$^{3}$-AmbigCaps are presented in Table \ref{table:amcaps}. When compared with other baseline methods, we observe that our proposed \ourmethod{} method also achieves significant performance improvements in all translation directions. In Table \ref{table:amcaps}, we observe that our proposed method \ourmethod{} outperforms by $+4.8$ BLEU scores on average when the visual modality is used, which is larger than that in M$^{3}$-Multi30K.


\subsection{Ablation Study}
In this section, we conduct comprehensive ablation study to demonstrate the effectiveness of different components in our proposed \ourmethod{} method on the test set of M$^{3}$-Multi30K.

\paragraph{Effect of LVPG.}
In Table~\ref{table:multi30k} and Table~\ref{table:amcaps},
we observe that our language-aware visual prompt generation (LVPG) brings significant improvements for Multilingual MMT. To demonstrate the effectiveness of LVPG, we further propose two alternative methods (i.e., \ourmethod{} (Static) and \ourmethod{} (CoCoOp)) to generate the visual prompts in Table~\ref{table:language-aware_generation}. 
Specifically, in \ourmethod{} (Static), we directly generate visual prompts by mapping the visual tokens $\{v_{m}\}_{m=1}^{M}$ using the mapping network, where the parameters of the mapping network are static after training and not conditioned on the target language token embedding $t_{L_j}$.
In Table \ref{table:language-aware_generation}, we observe that our \ourmethod{} outperforms these alternative methods a lot, which guides the visual clues to bridge the semantic gap between multiple languages.


\paragraph{Effect of Different Vision Backbones.}
In Table~\ref{table:backbone},
we compare the results of \ourmethod{} by using the visual tokens extracted by different vision backbones~\cite{he2016deep,dosovitskiy2020image} in CLIP. 
In Table \ref{table:backbone}, we observe that our \ourmethod{} achieves best results when using ViT-L/14 as the vision encoder.
Thus, we use  ViT-L/14 as the vision encoder by default.
Moreover,
we observe that the performance is better when the capacity of the vision backbone is better.
It is also reasonable because the quality of the visual tokens is better when using more powerful vision backbones.

\begin{figure*}[t]
\begin{center}
\includegraphics[width=0.85\linewidth]{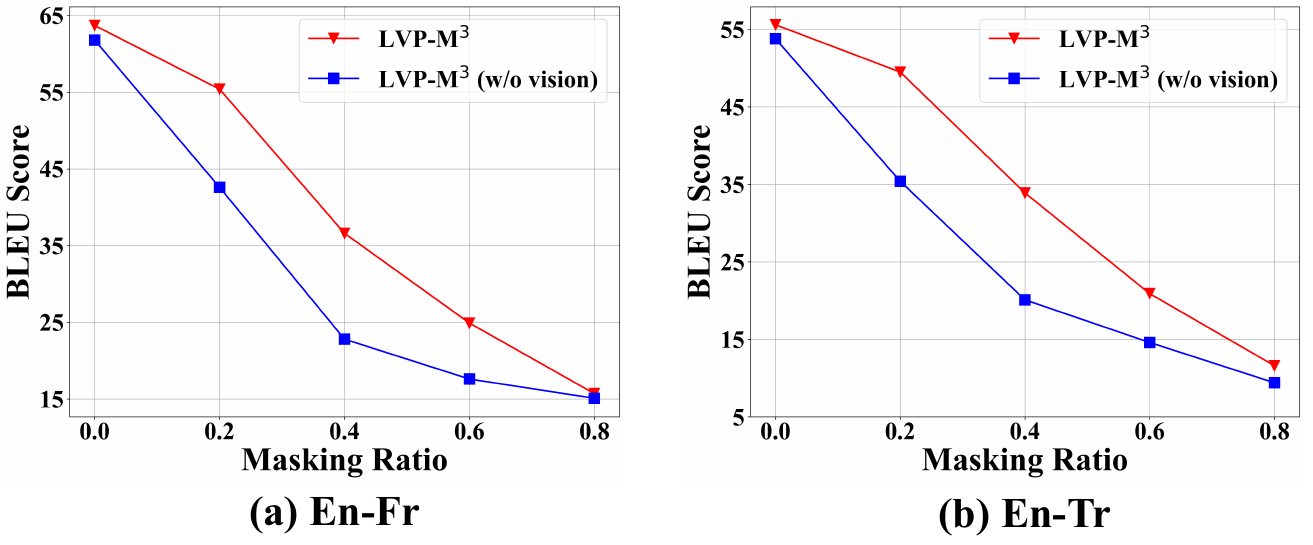}
\caption{Translation results of \ourmethod{} under different masking ratios on the source language. Results are evaluated on the M$^{3}$-Multi30K test set by translating English (En) to other languages (i.e., Fr and Tr).}
\label{fig:rateresults}
\end{center}
\end{figure*}

\subsection{Further Analysis}

\begin{figure*}[t]
\begin{center}
\includegraphics[width=1.0\linewidth]{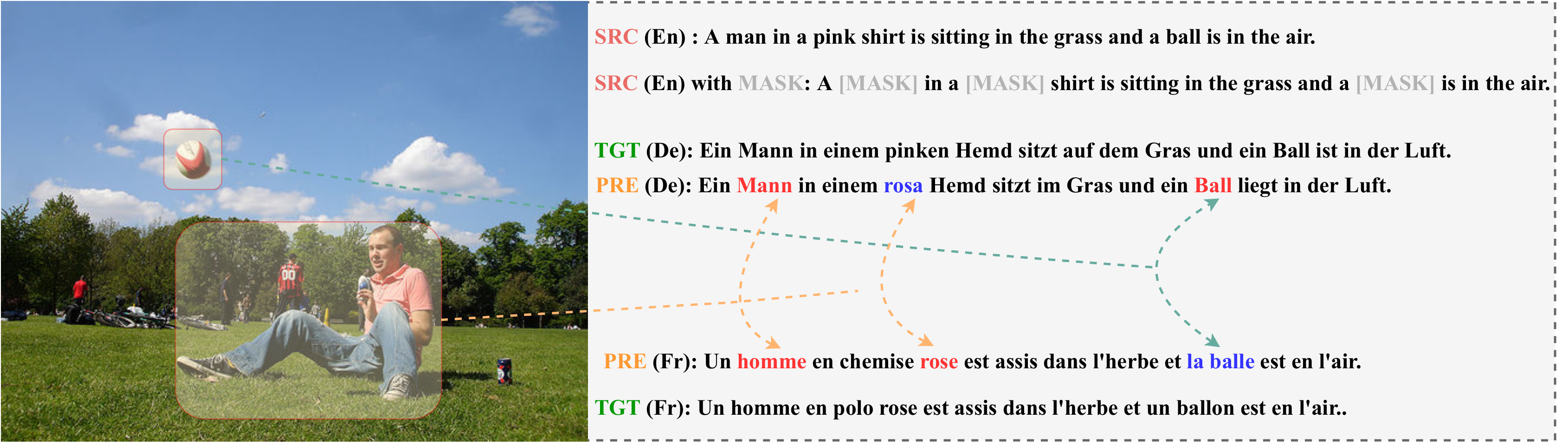}
\caption{A qualitative example by translating English (En) to German (De) and French (Fr) with the help of vision modality.
Tokens in red denotes correct translation.
Tokens in blue denotes good synonyms,
which have the similar meaning with the ground-truth of target language. \textbf{SRC} denotes the source language. \textbf{MASK} means the masked contents in the source language. \textbf{PRE} and \textbf{TGT} represent the predicted translation results and  the ground-truth of the target language, respectively.}
\label{fig:mask-visual}
\end{center}
\end{figure*}
\paragraph{Visualization of Different Masking Ratios.}
As shown in Fig.~\ref{fig:rateresults}, we compare our \ourmethod{} method with the alternative method  \ourmethod{} (w/o vision) to analyze the effectiveness of visual context when using different masking ratios on the source language. Specifically, in \ourmethod{} (w/o vision), we only use the Transformer encoder to process the source language with the target language embedding and then adopt the Transformer decoder to predict the target language for multilingual MT, where the vision encoder and LVPG are both not used in \ourmethod{} (w/o vision).

In Fig.~\ref{fig:rateresults},
we report the results of these methods by translating from English (En) to French (Fr) and Turkish (Tr).
First, when the ratio of masking increases, BLEU scores drop whether the visual contents are added or not,
and our \ourmethod{} still outperforms \ourmethod{} (w/o vision) a lot.
Second,
the performance gap between \ourmethod{} and \ourmethod{} (w/o vision) is larger
when the mask ratio is between 20\% and 40\%,
which shows that the visual information improves the robustness of the translation model.
Third, when the mask ratio is larger,
the results of these methods on all settings degrade.
When the mask ratio is set as 80\%, the results of \ourmethod{} (w/o vision) are close to those of \ourmethod{}.
It is also reasonable,
because most tokens in each source language are masked and it is difficult to translate well for both methods under these extreme scenarios. 


\paragraph{Qualitative Analysis.}
To further explore the necessity of visual modality for machine translation,
we compare the predictions results (i.e., De and Fr) of a sample source language (i.e., En) with the ground truth of these target languages in Fig.~\ref{fig:mask-visual}.
Specifically,
the ``man'' (noun), ``pink'' (adjective), and ``ball'' (noun) are masked,
and these masked tokens describe the saliency regions in the corresponding left image.
We have the following observations.
First, we observe that even though the ``man'' is masked,
the prediction results of German and French on this token are still right,
which means that visual modality is complementary rather than redundant if the text is insufficient.
Second, our model translates some tokens to their synonyms in the target language. For example, although the translated word {``rosa''} in German is evaluated as a bad translation for the masked token ``pink'' in English, it represents the same meaning as the word {``pinken''} in German. 
Besides, ``la balle'' in French is also the synonym of ``ball'' in English,
which further demonstrates the effectiveness of the vision modality.

\subsection{Discussion on ~\ourmethod{}}
 In our proposed ~\ourmethod{} method, first, both encoders (vision and text) and decoder are shared for all language pairs, while previous methods on MMT usually adopt different models for different language pairs. Second, to generate different visual prompts for different language pairs with minimal additional parameters, we just use controller network to generate the parameters of mapping network to map the vision embeddings. Third, different language translation directions are used in training, where the target language token is also prefixed to each source sentence for denoting the translation direction. Last, training separated models will result in huge training costs when compared with the multilingual models as discussed in many multilingual methods. 
 
\section{Conclusion}
In our work, we first propose the Multilingual MMT task to support the multilingual multimodal machine translations between different language pairs using one single model.
Then,
we propose an effective  LVP-M$^{3}$ baseline method for the Multilingual MMT task,
where a language-aware prompt generation module is proposed to generate visual prompts for different target languages dynamically.
Comprehensive experimental results on our established Multilingual MMT benchmark datasets demonstrate the effectiveness of our proposed  LVP-M$^{3}$ method for Multilingual MMT.
\section{Limitations}
Although our proposed  \ourmethod{} method has achieved substantial improvements for Multilingual MMT,
we find that there still exists some hyper-parameters (e.g., the number of encoder and decoder layers,) to tune for better results, which may be time-consuming. Besides, in our established datasets, we support seven languages currently,
and we will extend to support more languages and more translation directions for Multilingual MMT in the future work. 
\section*{Acknowledgments}
This work was supported in part by the National Natural Science Foundation of China (Grant Nos. 62276017, U1636211, 61672081), the 2022 Tencent Big Travel Rhino-Bird Special Research Program, and the Fund of the State Key Laboratory of Software Development Environment (Grant No. SKLSDE-2021ZX-18). 

\bibliography{anthology,custom}
\bibliographystyle{acl_natbib}

\clearpage



\end{document}